\documentclass{article}
\usepackage{ijcai18}

\usepackage{times}
\usepackage{xcolor}
\usepackage{soul}
\usepackage[utf8]{inputenc}
\usepackage[small]{caption}
\usepackage{subfigure}
\usepackage{graphicx}
\usepackage{amsfonts}
\usepackage{booktabs}
\usepackage{multirow}
\usepackage{mathtools}
\usepackage{caption}
\usepackage{balance}

\newlength\savedwidth

\title{\textsc{DeepTravel}: a Neural Network Based Travel Time Estimation Model \newline with Auxiliary Supervision }

\author{
	Hanyuan Zhang$^1$, 
	Hao Wu$^1$, 
	Weiwei Sun$^1$,
	Baihua Zheng$^2$
	\\ 
	$^1$ School of Computer Science, Fudan University, Shanghai, China \\
	$^2$ Singapore Management University, Singapore\\
	$^1$ \{hanyuanzhang16,wuhao5688,wwsun\}@fudan.edu.cn,
	$^2$ bhzheng@smu.edu.sg
}

\newenvironment{shrinkeqfront}[1]
{ \bgroup
	\addtolength\abovedisplayshortskip{#1}
	\addtolength\abovedisplayskip{#1}}
{\egroup\ignorespacesafterend}

\newenvironment{shrinkeqback}[1]
{ \bgroup
	\addtolength\belowdisplayshortskip{#1}
	\addtolength\belowdisplayskip{#1}}
{\egroup\ignorespacesafterend}

\begin{document}

\maketitle

\begin{abstract}

Estimating the travel time of a path is of great importance to smart urban mobility. Existing approaches are either based on estimating the time cost of each road segment which are not able to capture many cross-segment complex factors, or designed heuristically in a non-learning-based way which fail to utilize the existing abundant temporal labels of the data, i.e., the time stamp of each trajectory point. In this paper, we leverage on new development of deep neural networks and propose a novel auxiliary supervision model, namely \textsc{\small DeepTravel}, that can automatically and effectively extract different features, as well as make full use of the temporal labels of the trajectory data. We have conducted comprehensive experiments on real datasets to demonstrate the out-performance of \textsc{\small DeepTravel} over existing approaches.

\end{abstract}

\section{Introduction}
\label{sec:intro}



The advances in GPS-enabled mobile devices and pervasive computing techniques have generated massive trajectory data. The large amount of trajectory data provide opportunities to further enhance urban transportation systems. Estimating the travel time of a path at a certain time is an essential piece of information commuters desire to have. It is not a trivial problem as the travel time will be affected by many dynamics, such as the dynamics of the traffic, the dynamics at the crossroads, the dynamics of the driving behavior and the dynamics of the travel time of same paths in the historical data. These dynamics make the travel time indeterminate and hard to be estimated.



Existing solutions all adopt divide-and-conquer approach to perform the estimation by decomposing a path into a sequence of \emph{segments} or \emph{sub-paths}. Segment-based approaches~\cite{de2008traffic,asif2014spatiotemporal,lv2015traffic} estimate the travel time of each road segment individually, while the additional time spent at the intersection of segments due to traffic lights and turns is not considered. Moreover, they depend on high quality travel speed estimations/measurements, while the estimated speed is not accurate because of the sampling rate and GPS error. Consequently, the error of the estimated travel time will be accumulated along each road segment.
Sub-path based approaches~\cite{rahmani2013route,wang2014travel} try to estimate the time of the whole path by extracting the time consumption of sub-paths occurred in the historical dataset.
In general, sub-path based approaches perform better than segment-based ones. They can eliminate some errors accumulated by those segment-based approaches. However, they are still designed in an empirical and heuristic way but not training-based, which leaves the room for further improvement.

In summary, the main reason why existing estimation approaches could not achieve excellent accuracy is two-fold. They do not consider the path as a whole and they do not fully leverage the natural supervised labels of the data, i.e., the time stamp of each GPS sampling point that is easy to collect.  
On the other hand, thanks to the recent boom of deep learning researches, more problems can be solved by end-to-end models which significantly outperform the traditional heuristic approaches. Moreover, deep learning models have a strong representation power which enables the capturing of more latent features and the modeling of such complicated dynamics in travel time estimation problem.
 
Motivated by this, we propose a deep model named \textsc{\small DeepTravel} which can learn directly from the historical trajectories to estimate the travel time. \textsc{\small DeepTravel} is specifically designed through considering the characteristics of trajectory data by applying a new loss function for auxiliary supervision and is able to extract multiple features that affect the travel time. 
 As a summary,  our main contributions are as follows:
 
 \begin{itemize}
 	\vspace{-0.05in}
 	\item We propose \textsc{\small DeepTravel}, an end-to-end training-based model which can learn from the historical dataset to predict the travel time of a whole path directly. We introduce a dual interval loss to \emph{fully} leverage the temporal labeling information of the trajectory data which works as an auxiliary supervision.
 	\vspace{-0.05in}
 	\item We propose a feature extraction structure to extract features including spatial and temporal embeddings, driving state features, short-term and long-term traffic features. This structure can effectively capture different dynamics for estimating the travel time accurately.
 	\vspace{-0.05in}
 	\item  We conduct comprehensive experiments to evaluate our model with two real datasets. The results demonstrate the advantage of our model over the state-of-the-art competitors.
 \end{itemize}
 %

\section{Related Work}
\noindent

As stated in Section~\ref{sec:intro}, existing approaches on estimating the path travel time could be categorized into two clusters, \emph{segment-based} approaches and \emph{sub-path-based} approaches. The former one tries to estimate the travel time of each individual road segment in the network via different methods, e.g., the loop detectors~\cite{jia2001pems,rice2004simple}, support vector regression~\cite{asif2014spatiotemporal} and stacked autoencoder~\cite{lv2015traffic}. Approaches falling within this cluster are designed for estimating the travel time of a single road segment so they could not achieve a high accuracy when predicting the travel time of paths. The inaccuracy of the estimation is mainly caused by not considering the interaction between road segments. In addition, the estimation heavily depends on high quality travel speed data of each segment which might not be always available.

%

In order to overcome the weakness of the individual road segment-based methods,  sub-paths based approaches are proposed. They consider sub-paths instead of single segments as a way to include the interaction between road segments into the estimation. For example, \cite{han2011data,luo2013finding} mine frequent trajectory patterns; \cite{rahmani2013route} introduces a non-parametric method and utilizes the travel time of the common sub-paths between the query path and historical paths to estimate the travel time of the whole path after incorporating a list of potential biases corrections; \cite{wang2014travel} finds the optimal concatenation of trajectories for an estimation through a dynamic programming solution. They are able to improve the performance, as compared with segment-based approaches. However, the improvement is still limited due to the heuristical design, i.e., optimizing the error of the travel time is not the target.



On the other hand, deep learning methods have shown great power in modeling trajectory problems recently. For example, \cite{DBLP:conf/ijcai/SongKS16} uses recurrent neural network(RNN) to predict people’s future transportation mode in large-scale transportation networks; \cite{ijcai2017-430} models trajectory data with RNN, which can well capture long-term dependencies and achieve a better performance in predicting next movement than shallow models; \cite{ijcai2017-234} uses RNN with embeddings to represent the underlying semantics of user mobility patterns. Since RNN is suitable for modeling trajectory related problems, we will leverage on the power of RNN to perform travel time estimation of paths in this work.

\vspace{-0.1in}
\section{Problem Definition}

\begin{figure}
 	\vspace{-0.35in}
 	\centering
 	\includegraphics[scale=0.45]{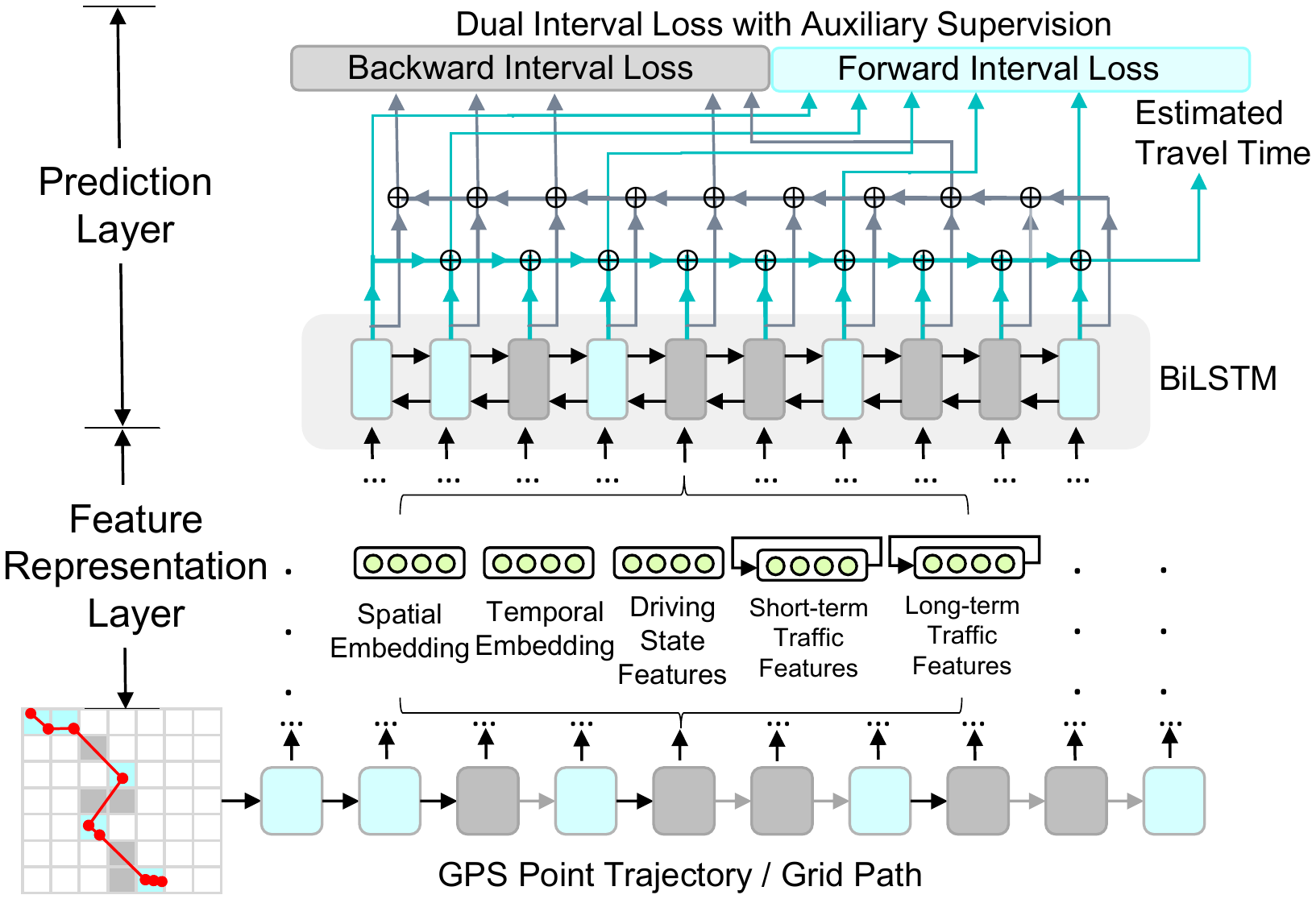}
 	\vspace{-0.1in}
 	\caption{The framework of \textsc{\small DeepTravel}. The trajectory are transformed into grid sequence. Grids having GPS points located in are in blue and others are in gray. $\oplus$ is the element-wise addition.}
 	\vspace{-0.15in}
 	\label{fig:global feature}
 \end{figure}

To adopt neural networks in our study and similar to many existing approaches~\cite{de2015artificial,zhang2017deep}, we partition the whole road network into $N \times N$ disjoint but equal-sized grids. Accordingly, a travel path $G$ started at $t_1$ could be represented by a sequence of grids it passes by, i.e., $G=\{g_1,g_2,...,g_n\}$. As long as the granularity of grid cells is fine enough, $G$ is able to capture the real movement of the path in road networks. Meanwhile, we assume sampled GPS points of the path are recorded to capture the real trajectory $T$ of $G$ in the form of $T=\{p_1, p_2, \cdots, p_m\}$. Each GPS point $p_i = (x_i,y_i,t_i)$ has latitude $x_i$, longitude $y_i$ and time stamp $t_i$, and the value of ($t_m-t_1$) indicates the real travel time of $T$. We can map a trajectory $T$ to a path $G$. Note some of the grids in $G$ will have one or multiple GPS points, while other grids might not have any, e.g., gray grids shown in Figure~\ref{fig:global feature}. We need to keep the grids with no GPS points to guarantee the continuity of a path. Our target is to use historical paths to train the model which can predict the travel time for a given path $G'$ that starts its travel at $t_1$.

\vspace{-0.1in}
\section{Solution}

In this section, we present our solution, i.e., the model \textsc{\small DeepTravel}. Figure~\ref{fig:global feature} shows that \textsc{\small DeepTravel} consists of two layers, the \emph{feature representation layer} and the \emph{prediction layer}. The former aims at extracting different features from the path, and the latter uses these feature representations to predict the travel time under auxiliary supervisions.

\vspace{-0.1in} 
\subsection{Feature Representation Layer}\label{sec:feature_extraction_layer}

We use features to capture the factors that could affect the travel time of paths. \textsc{\small DeepTravel} considers spatial and temporal embedding, driving state features, as well as short-term and long-term traffic features. We employ each grid as the carrier of these features. 
Note that we will study the effects of these features in the experiments.
%



\noindent
\textbf{Spatial and temporal embedding.} 
Both the spatial factor and the temporal factor affect the moving speed and hence the travel time. For example, the speed limits of different regions vary (e.g., residential areas and industrial districts usually have different speed limits); the traffic condition varies from time to time (e.g., traffic in peak hours is much heavier than that in non-peak hours), and also varies from place to place (e.g., 80\% of the car movements only pass by 20\% of the road segments and hence certain regions have a much higher possibility to encounter traffic jam). 
However, capturing all these factors precisely is not an easy task. We purposely train our model \textsc{\small DeepTravel} to learn the characteristics w.r.t. each grid automatically. In order to achieve this goal, we adopt the distributed representation to represent each grid using a low-dimensional vector $V \in \mathbb{R}^d$. The distributed representation has been widely used as a representation learning method, such as Word2Vec in natural language  ~\cite{mikolov2013distributed}, and deepwalk~\cite{perozzi2014deepwalk} in social networks. The spatial embedding vector $V_{sp}$ can contain a variety of feature information of the grid, which is scattered in various bits. Similar as spatial embeddings, we use distributed representation to represent temporal features. We divide the day into different time-bins (e.g. an hour a bin in our experiments), and use an unique vector $V_{tp}$ to represent each time-bin. Both $V_{sp}$ and $V_{tp}$ could be initialized randomly, and updated during the training of the model. 
\noindent
\textbf{Driving state features.} 
The driving process of vehicles can often be divided into the \emph{starting stage}, the \emph{middle stage} and the \emph{ending stage}, and vehicles have different driving characteristics in various stages. For example, a vehicle prefers driving on the main roads/highways in the middle stage, where the speed could be very fast; while it has to move from the source of the journey to the main roads/highways in the starting stage and it has to move from the main roads/highways to the destination in the ending stage. We use the vector $V_{dri} \in \mathbb{R}^4$ to represent the driving state features. It contains three 0-1 bits which represent the starting, middle and ending stages respectively and a ratio value capturing the proportion of the current path that is traveled (e.g., $[1, 0, 0, 0.2]$ indicates a starting stage and it finishes $20\%$ of the entire path).

\noindent
\textbf{Short-term and long-term traffic features.} 
Traffic condition in a sub-region has the characteristic of continuity in terms of time dimension, e.g., a road segment that experienced traffic jam from 8:00 to 8:30 this morning is expected to have heavy traffic at 8:35, which means the traffic condition of the path right before a query is issued on the path is informative and useful. Accordingly, we use the term $V_{short}$ to represent the short-term traffic condition features. 

\begin{figure}
	\vspace{-0.25in}
	\centering
	\includegraphics[scale=0.3]{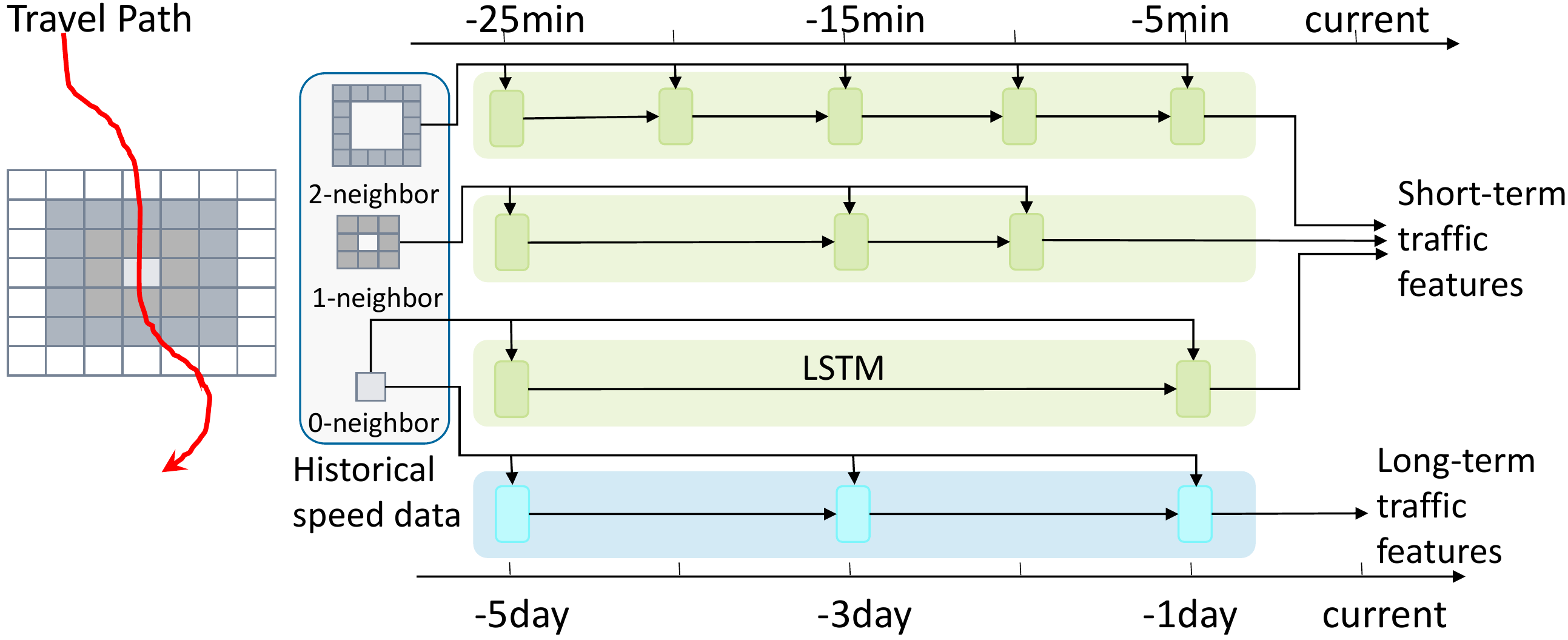}
	\vspace{-0.1in}
	\caption{The short-term and long-term traffic feature extraction.}
	\vspace{-0.2in}
	\label{fig:speed_features}
\end{figure}

Given a query submitted at time $t$, we extract $V_{short}$ from historical trajectories falling within the time window of $[t-1 \text{ hour}, t)$. To be more specific, we partition trajectories into disjoint time-bin $\tau$s of $\delta$ minutes (e.g., 5 minutes in our experiment). Then, the traffic conditions of a certain grid $g_i$ along these short time-bins form a sequence which reflects the temporal evolvement of the traffic condition in $g_i$. 
Hence, we utilize the \emph{long short-term memory network} (LSTM) \cite{hochreiter1997long}, a typical recurrent neural network for sequence modeling, to capture such temporal dynamics. The LSTM is fed by sequences of the statistical information of each time-bin, e.g., $\tau_1 \sim \tau_{12}$, and we set $V_{short}$ to the last hidden state of LSTM. Notice that after partitioning the historical data into 5-minute-span time-bins, some grids may have no vehicle passing by in some time-bins. As LSTM model can handle variable length sequences, we can easily tackle this problem by \emph{skipping} those time-bins with no vehicle passing by. E.g., in Figure~\ref{fig:speed_features}, for the grid of "0-neighbor", only -5-minute and -25-minute time-bin have historical vehicles passing by,  while we can skip the remaining empty time-bins when feeding data into LSTM. Then, we design the input w.r.t. the $j$-th time-bin $\tau_j$ of grid $g_i$ in the form of 
\begin{shrinkeqfront}{-0.05in}
\begin{shrinkeqback}{-0.05in}
\begin{equation}
		{x}_{i}^j =\left(j, v_{j}, n_{j}, {len_{i}}/{v_{j}}\right)\label{eqn:traffic_input}
\end{equation}
\end{shrinkeqback}
\end{shrinkeqfront}

We include $j$ to indicate the degree of closeness to the current query time in a linear scale, i.e., $j=12$ infers that the time-bin is one hour before current time which has the least closeness, and $j = 1$ infers 5 minutes before, which has the largest closeness. $v_{j}$ is the mean speed estimated from the samples in $g_i$ at $\tau_j$; $n_{j}$ refers to the number of historical samples, which indicates the degree of trustworthiness (the larger the better) about the estimated speed $v_{j}$ as $v_{j}$ tends to be vulnerable to outliers if $n_j$ is very small. $len_{i}$ is the length of the query path $G$ overlapped with grid $g_i$, and ${len_{i}}/{v_{j}}$ is a rough estimation of the average travel time of the path $G$ spent within $g_i$. 

As mentioned before, the historical number of samples extracted at one day in a short time interval is not large which may result in data sparsity issue. Noticing the fact of spatial locality of the traffic condition, i.e., traffic conditions tend to be similar in adjacent grids, we further include the traffic feature of $g_i$'s neighbors' as a solution. 
The $d$-neighbor set ${\cal N}_d^{g_i}$ of grid $g_i$ is defined as the set of grid cells with their distances to $g_i$ being $d$, i.e., 
\begin{shrinkeqfront}{-0.05in}
\begin{shrinkeqback}{-0.05in}
\[{\cal N}_d^{g_i} = \{g_j\mid \max(|g_i.x - g_j.x|,|g_i.y - g_j.y|) = d \}\]
\end{shrinkeqback}
\end{shrinkeqfront}
where $g_l.x,g_l.y$ indicate the position of $g_l$ (e.g., ($g_i.x$, $g_i.y$)$=$($1$,$2$) denotes the $1^{st}$ row and $2^{nd}$ column in $N\times N$ grids). Accordingly, ${\cal N}_0$ contains $g_i$ itself, ${\cal N}_1$ consists of all the grid cells adjacent to $g_i$, and so on. The final short-term traffic feature of $g_i$ is the concatenation of $g_i$'s $d$-neighbor sets' short-term traffic features. Figure~\ref{fig:speed_features} shows an example of $d = 0, 1, 2$.

Previous work has shown that for estimating travel time, we should also learn the long-term traffic dynamics \cite{wang2014travel}. We can easily modify the above short-term traffic feature extraction structure for supporting long-term traffic feature $V_{long}$. In detail, we
construct the sequence along the dimension of \emph{days}, e.g., we use the statistical information like Eq.~(\ref{eqn:traffic_input}) for the grid at the same time but in previous 7 days.

\vspace{-0.1in}
\subsection{Prediction Layer}\label{sec:prediction_layer}

The prediction layer consists of two parts, namely \emph{BiLSTM} and \emph{dual loss}. The former is to combine feature representations of each grid to infer travel time information in hidden state vectors; while the latter is to further optimize the model. 

\vspace{0.05in}\noindent
\textbf{BiLSTM.} 
As compared with LSTM, bidirectional LSTM (BiLSTM)~\cite{graves2005framewise} utilizes additional backward information and thus enhances the memory capability. In our problem setting, we use BiLSTM to capture the information of every grid $g_i$ in the path from the starting point to $g_i$ and from the ending point to $g_i$ simultaneously. We concatenate the features extracted in Section~\ref{sec:feature_extraction_layer} together to get the global feature vector $V$ of the grid, i.e., $V = [V_{sp},V_{tp},V_{dri},V_{short},V_{long}]$. 
We feed $V$ of the present grid to BiLSTM at each step and get the $i$-th hidden states $\overrightarrow{h_i}$ and $\overleftarrow{h_{i}}$ of the forward and backward layer respectively. We then concatenate these two states to get the $i$-th hidden state $h_i = [\overrightarrow{h_i},\overleftarrow{h_{i}}]$. 

 \begin{figure}
 	\vspace{-0.25in}
 	\centering
 	\includegraphics[scale=0.6]{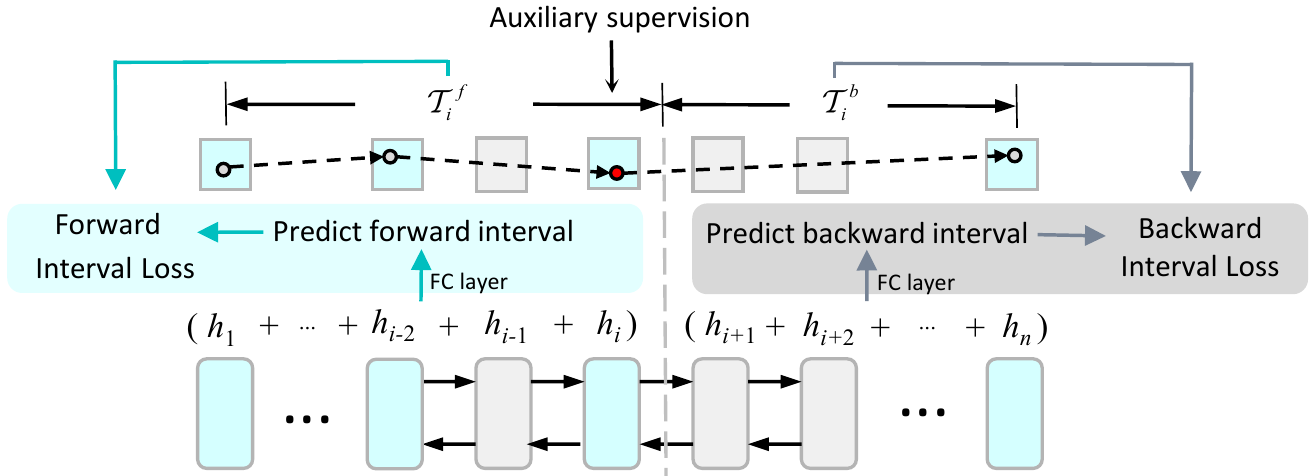}
 	\vspace{-0.15in}
 	\caption{The example of dual interval loss for auxiliary supervision.}
 	\vspace{-0.2in}
 	\label{fig:dual interval loss}
 \end{figure}

\vspace{0.05in}\noindent
\textbf{Dual interval loss for auxiliary supervision.} 
A simple way to estimate the travel time is to perform linear regression on the final hidden state $h_{n}$ by employing loss function such as mean squared error w.r.t. the ground truth $t_m-t_1$. Since this trivial loss does not utilize the intermediate time information from trajectory point time stamps, it wastes much useful information to supervise the model. To leverage such additional supervision information, we design a dual interval loss mechanism for auxiliary supervision which exactly matches the characteristic of BiLSTM.

The dual interval loss is constructed by two losses, the \emph{forward interval loss} and the \emph{backward interval loss}. The general idea is to force the model to learn to simultaneously predict the time interval from the start point to each intermediate GPS point $p_j$, i.e., the forward interval, and the interval from $p_j$ to the destination, i.e., the backward interval, as shown in Figure~\ref{fig:dual interval loss}. 
In detail, we construct forward/backward mask vector $\mathcal{M} \in \{0,1\}^n$ for activating forward/backward interval loss at some grids having supervisory information. Moreover, we construct $\mathcal{T}^f$, $\mathcal{T}^b \in {\mathbb R}^n$ for recording the forward and backward interval ground truth. The details can be found as follows.
\begin{shrinkeqfront}{-0.1in}
\begin{shrinkeqback}{-0.05in}
\begin{eqnarray}
\vspace{-0.15in}
\mathcal{M}_i &= &
\begin{cases}
1 & \text{if there is a point sampled in $g_i$} \\
0 & \text{otherwise} \\
\end{cases}\nonumber\\
\mathcal{T}^f_i &=&
\begin{cases}
g_i.t - g_{0}.t & \text{if there is a point sampled in $g_i$}\\
1 & \text{otherwise, a random value}  \\
\end{cases}\nonumber\\
\mathcal{T}^b_{i} &=&
\begin{cases}
g_n.t - g_{i}.t & \text{if there is a point sampled in $g_{i}$}\\
1 & \text{otherwise,  a random value}  \\
\end{cases}\nonumber  \nonumber
\end{eqnarray}
\end{shrinkeqback}
\end{shrinkeqfront}
\noindent
$g_i.t$ refers to the time when the vehicle \emph{leaves} the grid $g_i$ along the path, which can be derived from the corresponding trajectory data if there are GPS points sampled in the grid $g_i$. Specifically, we define $g_0.t$ as the time stamp of the first GPS point and $g_n.t$ is the time stamp of the last GPS point. 

For predicting the dual intervals, instead of using the current $h_i$ for prediction, we decide to adopt a \emph{summation} operation, which adds $h_1$ to $h_i$ together for forward prediction and $h_{i+1}$ to $h_n$ for backward prediction as Figure \ref{fig:dual interval loss} shows. 
The reason is that such summation operation feeds the model the prior knowledge of the summation property of time, i.e., the time spent on one path is the summation of the time spent on two sup-paths, and there is no need to learn such summation property from the data. Moreover, the predicted time of each step, i.e., $W^\top h_i +b$, now represents the time spent \emph{only} on grid $g_i$ which forces the sum of the predicted forward interval and the backward interval hold the same across all steps, i.e., ($W^\top\sum_{i=1}^n h_i +b$), which is exactly the travel time of the whole path. Thus, minimizing the dual loss can also benefit estimating the travel time of the entire path at each step which can be naturally regarded as the cheif supervision in our task.
Consequently, the forward/backward interval time estimation vectors $\mathcal{\hat {\cal T}}^f, \mathcal{\hat {\cal T}}^b$ are as follows.
\begin{shrinkeqfront}{-0.1in}
\begin{shrinkeqback}{-0.05in}
\begin{eqnarray}
\vspace{-0.15in}
\mathcal{\hat {\cal T}}^f&=&W^\top \left[h_1,h_1+h_2,...,\sum_{i=1}^{n-1}h_i,\sum_{i=1}^{n}h_i\right]+b \nonumber\\
\mathcal{\hat {\cal T}}^b&=&\left[W^\top \left[\sum_{i=2}^{n}h_i,\sum_{i=3}^{n}h_i,...,h_{n-1}+h_{n},h_{n}\right]+b, 0 \right]  \nonumber
\vspace{-0.1in}
\end{eqnarray}
\end{shrinkeqback}
\end{shrinkeqfront}
Here, $\mathcal{\hat {\cal T}}^f\in \mathbb{R}^n$ represents the travel time from the starting point to each grid in the path, and $\mathcal{\hat {\cal T}}^b \in \mathbb{R}^n$ represents the travel time from each grid in the path to the ending point. 
For both forward and backward predictions, we use the shared weight $W, b$ because we want to restrict the task of transformation from $h_i$ to the travel time spent on grid $g_i$ to be the same in both forward and backward predictions. 
The dual interval loss is the summation of the forward and backward interval losses. We use the relative mean square error $\mathcal{L}$ as follows. Note, operation with $``[~]"$ indicates the element-wise one.
\begin{shrinkeqfront}{-0.1in}
\begin{shrinkeqback}{-0.05in}
\begin{equation}\notag
\begin{footnotesize}
\mathcal{L} = \frac{
	\mathcal{M} ^ \top \cdot 
    \left( 
		     (\mathcal{{\hat T}}^f-\mathcal{T}^f) [/] {\mathcal{T}^f}
    \right)^{[2]} + 
	\mathcal{M} ^ \top \cdot 
	\left( 
	(\mathcal{{\hat T}}^b-\mathcal{T}^b) [/] {\mathcal{T}^b}
	\right)^{[2]}
}
{
	\textbf{1}^\top \cdot (\mathcal{M}[*]2)
} 
\end{footnotesize}
\end{equation}
\end{shrinkeqback}
\end{shrinkeqfront}
The dual interval loss not only minimizes the travel time estimation error of the whole path but also constrains the forward and backward interval estimation error of intermediate grids, which utilizes the intermediate time information of a trajectory. It has the following three advantages. First, these intermediate monitoring information to some extent increases the amount of data to help model training better. Second, adding the supervisory information in the middle can make the loss signal back-propagate more accurate and effective, which will reduce the risk of vanishing gradient for long sequences. Third, the dual loss exactly matches the BiLSTM characteristics, as each step of BiLSTM has the information from the starting grid to the current grid and that from the current grid to the ending grid, which can naturally be used by forward and backward interval loss. We will show the superiority of the dual interval loss in the experiment section.

\vspace{-0.1in}
\subsection{Training}
The goal of \textsc{\small DeepTravel} is to minimize the dual loss function $\mathcal{L}$. In other words, denoting the trainable parameters in \textsc{\small DeepTravel} as $\theta$, and the spatial and temporal embedding vectors as $\mathcal{E}$, Eq.~(\ref{eqn:lost_f}) defines our goal. Here, $S$ is the number of training trajectories and $\mathcal{L}^{(i)}$ is the dual loss function of $i$-th trajectory data. The model is trained by employing the derivative of the loss w.r.t. all parameters through back-propagation-through-time algorithm~\cite{werbos1990backpropagation}.
\vspace{-0.1in}
\begin{equation}
\label{eqn:lost_f}
\min_{\theta,\mathcal{E}} \sum_{i=1}^{S} \mathcal{L}^{(i)}(\theta,\mathcal{E}) 
\end{equation}

\begin{table}
	\vspace{-0.3in}
	\caption{The description and statistics of the datasets.}
	\vspace{-0.15in}
	\label{data_stats}
	\makebox[\linewidth][c]{
		\begin{small}
			\begin{tabular}{c|c|c}
				\hline
				Dataset & Porto & Shanghai \\
				\hline
				trajectory number & 420,000 & 1,018,000 \\
				sampling interval & 15s & 10s \\
				area & $16,735\rm{m} \times 14,389m $ & $29,833\rm m \times 37,867m$\\
				grid size & $128 \times 128$ &  $256 \times 256$\\
				travel time mean & 762.60s & 954.59s \\
				travel time std  & 347.92s & 460.71s \\
				\hline	
			\end{tabular}
		\end{small}	
	}
	\vspace{-0.18in}
\end{table}

\begin{table*}
	\vspace{-0.3in}
	\caption{Performance comparison of \textsc{\small DeepTravel} and its competitors.}
	\label{big exp}
	\vspace{-0.15in}
	\centering
	\begin{footnotesize}	
		\begin{tabular}{c|c|ccc|ccc}
			\hline
			\multicolumn{2}{c}{Dataset} &\multicolumn{3}{|c|}{Porto}  & \multicolumn{3}{c}{Shanghai } \\
			\hline
			\multicolumn{2}{c}{Metrics} & \multicolumn{1}{|c}{\textit{MAE} (sec)} & \multicolumn{1}{c}{\textit{RMSE} (sec)}   & \multicolumn{1}{c}{\textit{MAPE}} & \multicolumn{1}{|c}{\textit{MAE} (sec)} & \multicolumn{1}{c}{\textit{RMSE} (sec)}  & \multicolumn{1}{c}{\textit{MAPE}}  \\
			\hline
			\multirow{5}{*}{Segment Based} & spd-MEAN & 245.87 & 358.32  & 0.2847 & 430.74 & 550.43 & 0.4170\\ 
			
			& ARIMA~\cite{ahmed1979analysis} & 227.40 & 517.51 & 0.2757 & 315.22 & 444.42 & 0.3074\\ 
			
			& SVR~\cite{asif2014spatiotemporal} & 241.41 & 353.35 & 0.2819 & 424.12 & 543.28 & 0.4085  \\
			
			& SAE~\cite{lv2015traffic} & 222.06 & 357.02  & 0.2734 & 310.47 & 413.62 & 0.3013 \\
			
			& spd-LSTM \cite{ma2015long} & 217.37 & 334.00 & 0.2624 & 302.45 & 397.48 & 0.2945 \\
			\hline
			\multirow{2}{*}{Sub-path Based} & RTTE~\cite{rahmani2013route} & 169.45  & 272.22 & 0.2234 & 214.01 & 307.77 & 0.2362 \\ 
			
			& PTTE~\cite{wang2014travel} & 159.43 & 268.11 & 0.2072 & 168.48 & 248.92 & 0.1914 \\
			\cline{2-8}
			\hline
			\multirow{4}{*}{End-to-End} & grid-MLP & 255.33 & 377.27 & 0.2933 & 423.53 & 541.19 & 0.3906 \\ 
			
			& grid-CNN & 250.86 & 363.17 & 0.2874 & 420.05 & 537.86 & 0.3885 \\
			
			& grid-LSTM & 180.27 & 300.98 & 0.2334 & 235.74 & 348.30 & 0.2463 \\
			
			
			& {\small \textsc{DeepTravel}} & \textbf{113.24} & \textbf{219.25} & \textbf{0.1337} & \textbf{126.59} & \textbf{196.85} & \textbf{0.1330} \\
			\hline 
		\end{tabular}
	\end{footnotesize}
	\vspace{-0.2in}
\end{table*}

\vspace{-0.15in}
\section{Experiments}

We conduct comprehensive experiments to compare the performance of {\small \textsc{DeepTravel}} and existing competitors. 
%
Source code and implementation details are available online at https://github.com/**anonymized for double-blind review**.

\vspace{-0.1in}
\subsection{Experiment Setting}

%
\textbf{Datasets.} 
Two real trajectory datasets are used in our experimental study, namely \textit{Porto} and \textit{Shanghai}.
%
%
The Porto dataset (http://www.kaggle.com/c/pkdd-15-predicttaxi-service-trajectory-i) is a 1.8GB open dataset, generated by 442 taxis from Jan. 07, 2013 to Jun. 30, 2014. The Shanghai one is generated by $13,650$ taxis from Apr. 01 to Apr. 17 in 2015 with the size of 16GB. We extract the trajectory trips occupied by passengers as valid trajectories. Table~\ref{data_stats} reports the description and statistics of the two datasets.


\noindent
\textbf{Hyperparameters.}
For the hyperparameters of our model, we split each dataset into training set, validation set and test set in the ratio of 8:1:1. 
%
The embedding size of spatial and temporal embeddings is set to 100 and initialized uniformly by [-1.0, 1.0]. We set the hidden unit as 100 for the both LSTM in traffic feature extraction and BiLSTM in prediction. We train the model using Adam algorithm~\cite{kingma2014adam} with an initial learning rate at 0.002. All the weights are uniformly initialized by [-0.05,0.05].

\noindent
\textbf{Metrics.}
We adopt \emph{mean absolute error} (MAE), \emph{mean absolute percentage error} (MAPE) and \emph{root-mean-squared error} (RMSE) as the major performance metrics, similar to existing approaches~\cite{rahmani2013route,wang2014travel}.

\noindent
\textbf{Approaches for comparison.} 
As mentioned before, existing approaches on estimating the path travel time are either segment-based or sub-path based. We implement \textit{spd-MEAN}, \textit{ARIMA}, \textit{SVR}, \textit{SAE}, \textit{spd-LSTM} as representatives of segments-based approaches, and \textit{RTTE} and \textit{PTTE} as representatives of sub-path based approaches. 
To be more specific, spd-MEAN estimates the speed of every segment by averaging from historical speeds. The remaining four segment-based approaches use different time series prediction models to predict the present speed of each segment given historical travel speeds, i.e., ARIMA uses auto-regressive integrated moving average model, SVR uses support vector regression model, SAE uses stacked auto-encoder model and spd-LSTM uses an LSTM model.
RTTE develops a non-parametric approach which uses the travel time of the common sub-paths between the query path and historical paths 
and PTTE finds the optimal concatenation of trajectories through a dynamic programming solution. To the best of our knowledge, PTTE is the best practice for the problem studied in this paper. 

In addition to the above seven existing competitors, we also propose three simple end-to-end models as baselines, namely \textit{grid-MLP}, \textit{grid-CNN} and \textit{grid-LSTM}. grid-MLP uses multilayer perceptron (MLP) model to predict the travel time of the path. We use a $N \times N$ matrix $M$ as the input, with each element $M_{ij}$ capturing the travel length that the vehicle passes through the grid $g_{ij}$; and we use two hidden layers with 1024 units and sigmoid as activation function. grid-CNN uses convolutional neural network (CNN) model to perform the estimation. It accepts the same input $M$ as grid-MLP. We use three convolutional layers and three max-pooling layers. Each convolutional layer has 64 $3\times3$ filters with stride 1; and each max-pooling is in the size of $2\times2$. Then it is followed by a fully-connected layer with 1024 units and sigmoid activation for prediction. grid-LSTM uses LSTM to predict the travel time. We set LSTM with 100 hidden units, and feed it with the travel length of the present grid at each step. All three models adopt the mean relative squared error as the loss function. Note that \textsc{\small DeepTravel} is also an end-to-end model.

%
%
%

\vspace{-0.10in}
\subsection{Overall Evaluation}
\vspace{-0.05in}
The first set of experiments is to evaluate the performance of estimation of the query path's travel time, with the results reported in Table~\ref{big exp}. We observe that in general the sub-path based approaches perform better than segment based approaches. This indicates that the interaction between adjacent road segments in a path is important.  
For segment based approaches, spd-LSTM outperforms others which demonstrates the power of LSTM model in capturing the features of time series data. For sub-path based approaches, PTTE performs better than RTTE since PTTE has an object function to model the trade-off between the length of a sub-path and the number of trajectories traversing the sub-path. For end-to-end approaches, \textsc{\small DeepTravel} is significantly better than others in all metrics. That is to say, a trivial neural network model can not predict the travel time well, and it is necessary to extract different features and adopt a more effective structure to construct the model like \textsc{\small DeepTravel} does. Note that grid-LSTM performs better than grid-MLP and grid-CNN. This is because a path only occupies a small part of grids in the whole city ($<1\%$). Accordingly, most elements of the input matrix $M$ are zero and hence grid-MLP and grid-CNN are not able to learn such valid features well. 

On the other hand, \textsc{\small DeepTravel} outperforms all the competitors with significant advantages. 
We can also observe from the results that segment-based approaches perform worse in Shanghai dataset than in Porto dataset; while sub-path based approaches and \textsc{\small DeepTravel} are more robust in different datasets. Based on our understanding of the datasets, trajectories in Porto are sparser but the traffic condition of Shanghai changes more drastically. The results demonstrate that \textsc{\small DeepTravel} works very well for the different challenges faced by different datasets.

\vspace{-0.1in}
\subsection{Performance of \textsc{DeepTravel}}
\noindent
\textbf{The impact of different features.} 
As {\small \textsc{DeepTravel}} takes in inputs from multiple features, we conduct the second set of experiments to study their effectiveness. We implement five different versions of {\small \textsc{DeepTravel}} with each taking in different feature inputs. \textit{ST} only uses the spatial and temporal embeddings; \textit{NaiveTraf} takes in the mean historical speed corresponding to the grid as the traffic feature; \textit{Traf} only uses the traffic features in our model; \textit{ST+Traf} accepts both traffic features as well as spatio-temporal embeddings as input; and \textit{ST+Traf+DS} takes in all the features considered by {\small \textsc{DeepTravel}} (DS refers to driving state feature).
%
%
%
As listed in Table~\ref{features exp}, {ST+Traf+DS} outperforms other versions. {ST+Traf} performs better than both {ST} and {Traf}, which means that both traffic features and spatio-temporal embeddings play important roles in the prediction. 
The driving state feature also improves the performance, as {ST+Traf+} performs better than {ST+Traf}. The result of {NaiveTraf} is not as good as that of {Traf} especially in Shanghai dataset, which means that our construction of traffic feature is more effective than trivially doing statistics, i.e., averaging historical speeds. It is worth noting that {ST} is better than {Traf} in Porto but worse than {Traf} in Shanghai, which showcases that the travel time of a path is greatly influenced by spatial location and time period in Porto, while it is mainly affected by the traffic condition in Shanghai which is a metropolis with heavy traffic flows.

 \begin{table}
	\vspace{-0.2in}
	\caption{Performance of  {\small \textsc{DeepTravel}} with different features.}
	\vspace{-0.15in}
	\label{features exp}
	\centering
	\begin{small}
	\begin{tabular}{c|cc|cc}
		\hline
		\multicolumn{1}{c}{Dataset} &\multicolumn{2}{|c|}{Porto} & \multicolumn{2}{|c}{Shanghai}  \\
		\hline
		\multicolumn{1}{c}{Metrics} & \multicolumn{1}{|c}{\textit{MAE} (sec)}   & \multicolumn{1}{c|}{\textit{MAPE}} &  \multicolumn{1}{|c}{\textit{MAE} (sec)}   & \multicolumn{1}{c}{\textit{MAPE}}   \\
		\hline
		ST & 129.33 & 0.1505 & 197.58  & 0.1926 \\ 
		
		NaiveTraf & 144.41 & 0.1688 & 199.06 & 0.1940 \\
		
		Traf & 132.28 & 0.1537 & 153.95 & 0.1559 \\
		
		ST+Traf & 114.47 & 0.1367 & 129.44 & 0.1362 \\
		
		ST+Traf+DS & \textbf{113.24} & \textbf{0.1337} & \textbf{126.59} & \textbf{0.1330}  \\
		\hline 
	\end{tabular}
	\end{small}
	\vspace{-0.1in}
\end{table}

 \begin{table}
 	\vspace{-0.05in}
 	\caption{The effectiveness of different loss functions.}
 	\label{different loss}
 	\vspace{-0.15in}
 	\centering
 	\begin{small}
 	\begin{tabular}{c|cc|cc}
 		\hline
 		\multicolumn{1}{c}{Dataset} &\multicolumn{2}{|c|}{Porto} & \multicolumn{2}{|c}{Shanghai}  \\
 		\hline
 		\multicolumn{1}{c}{Metrics} & \multicolumn{1}{|c}{\textit{MAE} (sec)}   & \multicolumn{1}{c|}{\textit{MAPE}} &  \multicolumn{1}{|c}{\textit{MAE} (sec)}   & \multicolumn{1}{c}{\textit{MAPE}}   \\
 		\hline
 		LSTM$_{no\_aux}$ & 130.57 & 0.1494 & 148.90  & 0.1506 \\ 
 		
 		BiLSTM$_{no\_aux}$ & 128.85 & 0.1476 & 143.72 & 0.1475 \\
 		
 		BiLSTM$_{for\_aux}$ & 115.64 & 0.1369 & 128.56 & 0.1349 \\
 		
 		BiLSTM$_{back\_aux}$ & 115.85 & 0.1372 & 128.77 & 0.1355 \\
 		
 		BiLSTM$_{dual\_aux}$  & \textbf{113.24} & \textbf{0.1337} & \textbf{126.59} & \textbf{0.1330}  \\
 		\hline 
 	\end{tabular}
 	\end{small}
 	\vspace{-0.23in}
 \end{table}

\noindent
\textbf{The effectiveness of different loss functions.} 
In order to demonstrate the effectiveness of the proposed dual interval loss with auxiliary supervision, we compare it with other loss functions. We construct five baselines which share the same feature extraction layer as {\small \textsc{DeepTravel}} but different loss functions for training. 
To be more specific, \textit{LSTM$_{no\_aux}$} feeds features to an LSTM, and only uses the \emph{final} hidden vector to predict the travel time (i.e., no auxiliary supervision) with the mean relative squared error for the loss. 
\textit{BiLSTM$_{no\_aux}$} is similar to \textit{LSTM$_{no\_aux}$}, i.e., uses the final forward and backward hidden state of BiLSTM for prediction. 
Both \textit{BiLSTM$_{for\_aux}$} and \textit{BiLSTM$_{back\_aux}$} leverage the auxiliary supervision, i.e., the time stamps of intermediate GPS points, but \textit{BiLSTM$_{for\_aux}$} only uses the forward interval loss as the loss function while \textit{BiLSTM$_{back\_aux}$ } only optimizes the backward loss.  
\textit{BiLSTM$_{dual\_aux}$} is \textsc{\small DeepTravel} model which optimizes both forward and backward interval loss functions with auxiliary supervision.

\begin{figure} 
		\vspace{-0.25in}
	\subfigure[Porto]{\label{fig:convergence_porto}
		\includegraphics[width=0.45\linewidth]{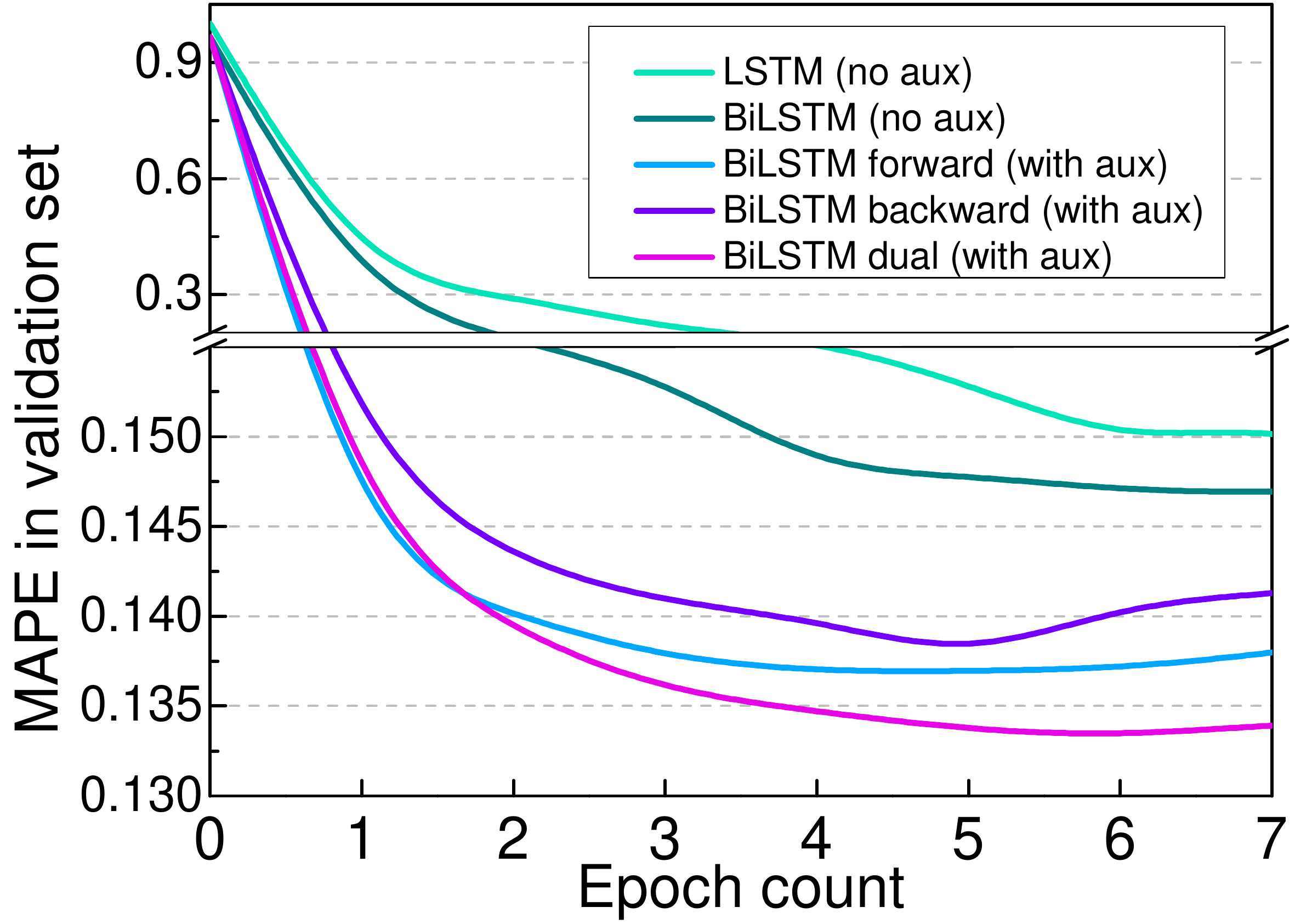}}
	\subfigure[Shanghai]{\label{fig:convergence_shanghai}
		\includegraphics[width=0.45\linewidth]{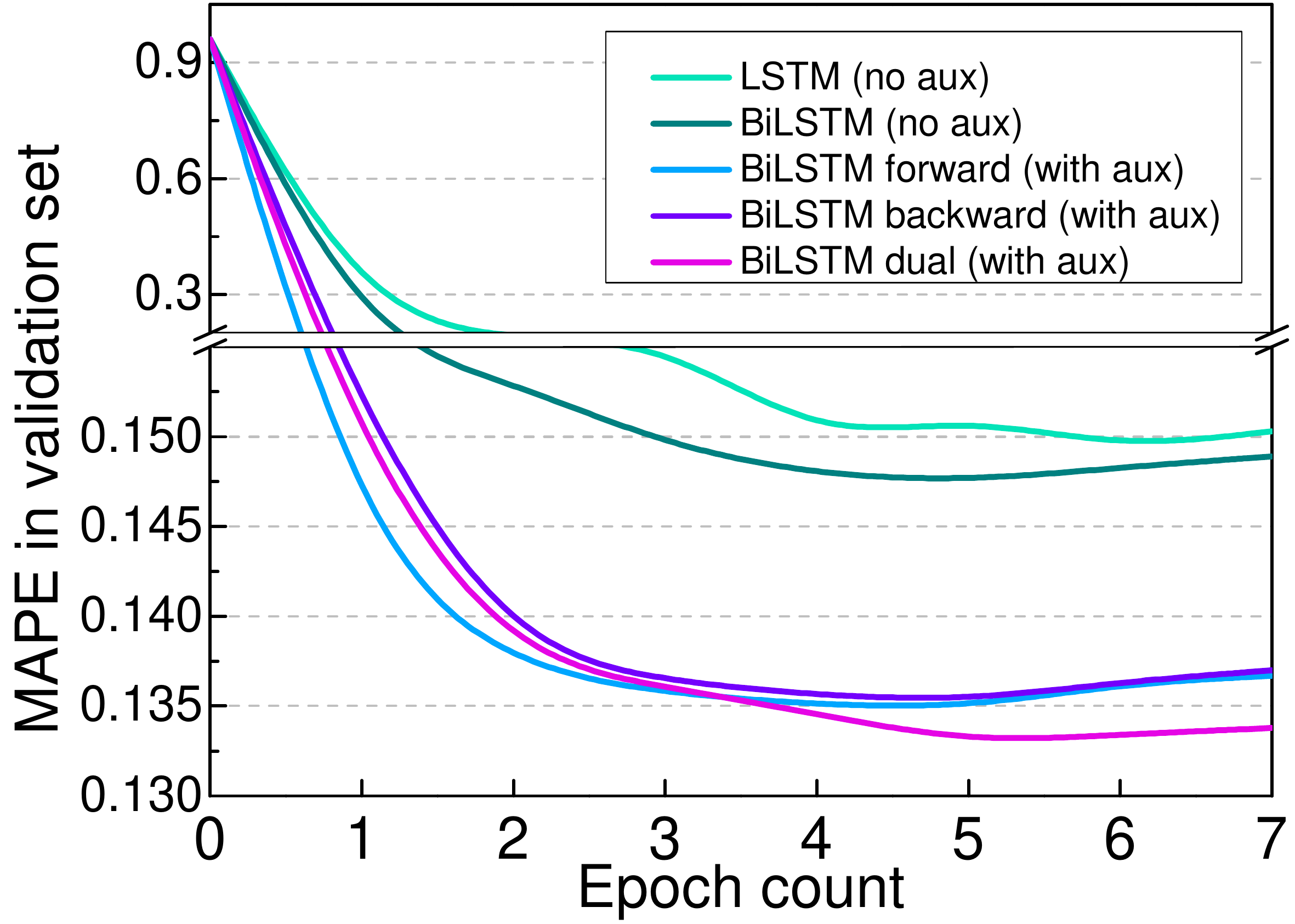}}
	\vspace{-0.2in}
	\caption{The MAPE curve under different loss functions.}
	\label{fig:convergence}
	\vspace{-0.15in}
\end{figure}

We report the quantitative results in Table~\ref{different loss} and the MAPE curve in validation set w.r.t. training epochs in Figure~\ref{fig:convergence}. 
From the results, we can find that BiLSTM$_{no\_aux}$ performs better than LSTM$_{no\_aux}$, which means that BiLSTM is able to capture correlations between grids much better than LSTM.  
We also observe that all the three models with auxiliary supervision behave much better than models without auxiliary supervision and have a very fast convergence. This proves that the auxiliary supervision from additional interval loss benefits the back-propagation of loss signals, and the additional supervision is some kind of data augmentation which can improve the results, as analyzed in Section~\ref{sec:prediction_layer}.
Last, as we expect, BiLSTM$_{dual\_aux}$ performs better than BiLSTM$_{for\_aux}$ and BiLSTM$_{back\_aux}$, which means auxiliary supervisions from forward and backward interval loss are not duplicate but complementary.

\begin{figure} 
	\subfigure[MAPE@Shanghai]{\label{fig:mape_shanghai}
		\includegraphics[width=0.45\linewidth]{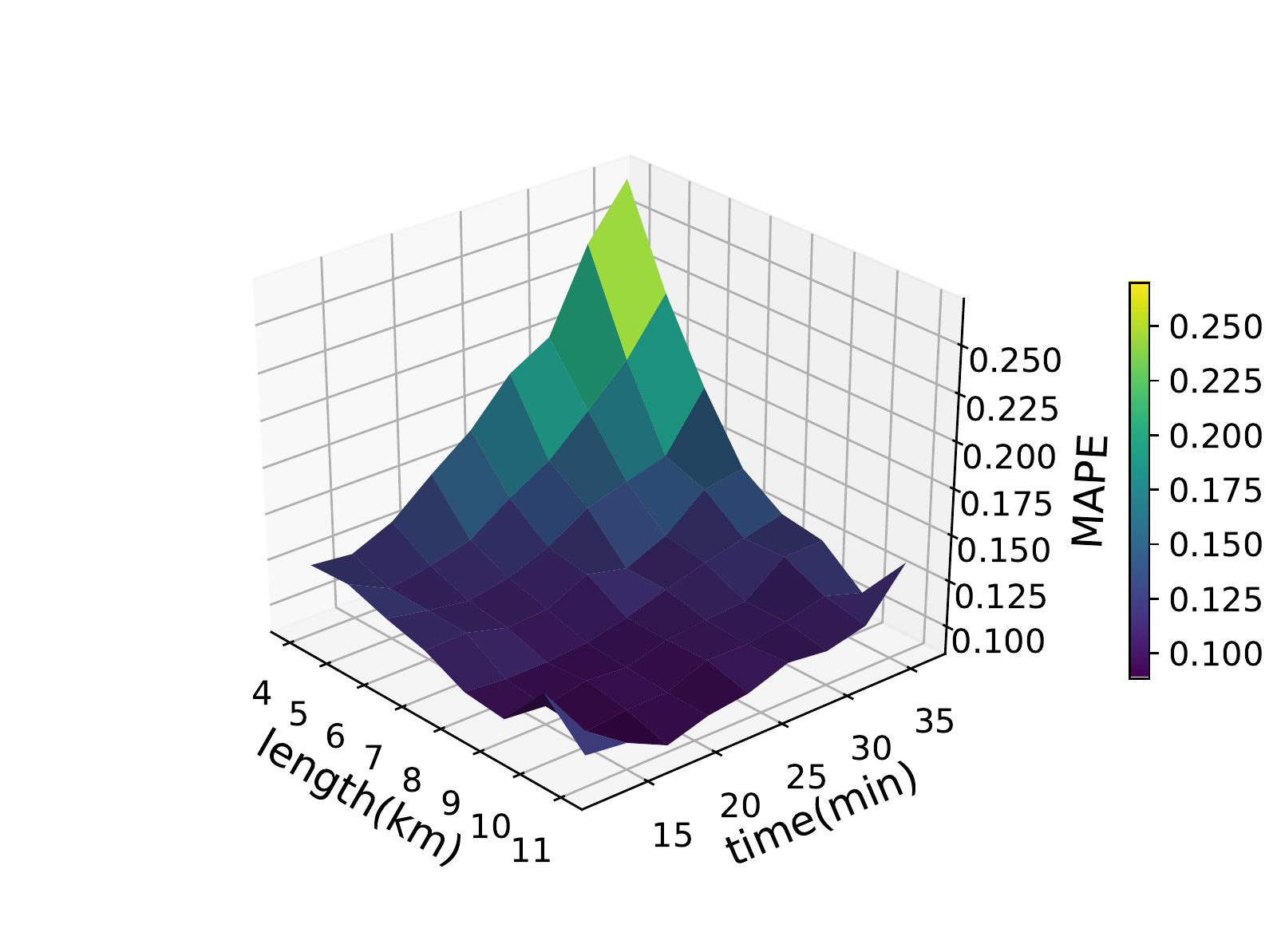}}
	\subfigure[Data frequency@Shanghai]{\label{fig:traj_hist}
		\includegraphics[width=0.45\linewidth]{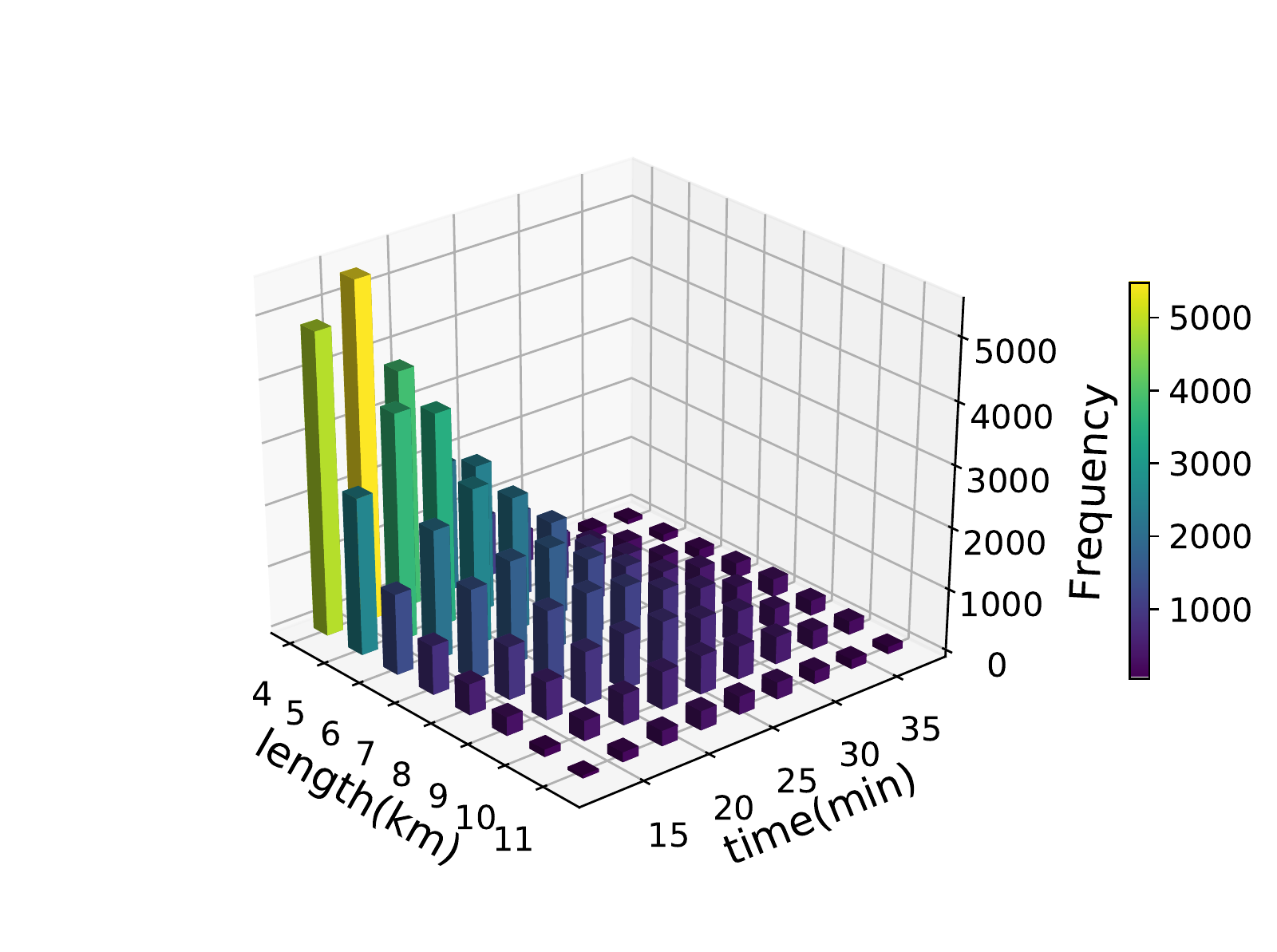}}
	\vspace{-0.2in}
	\caption{Performance of {\small \textsc{DeepTravel}} vs. length and travel time of paths.}
	\label{fig:length and time}
	\vspace{-0.2in}
\end{figure}

%
%
%
%
%

\vspace{0.03in}
\noindent
\textbf{The performance of  {\small \textsc{DeepTravel}} vs. length and travel time of paths.}
Last but not least, we partition the testing trajectory set into different subsets according to the length of the trajectories and the duration of the travel time, and report the MAPE of \textsc{\small DeepTravel} under Shanghai dataset, as a representative. The results are reported in Figure~\ref{fig:mape_shanghai}. 
In general, \textsc{\small DeepTravel} performs well (i.e.., MAPE around 0.1). However, we do observe a performance drop when the path is short and the travel time is long, e.g., 4km and 35min. Firstly, this type of trajectories is abnormal as the travel time in most cases is proportional to the length of the path. For example, the paths with the length of 4km and the travel time of 35min mean the average speed is about 6.9km/h which is extremely slow, only a little bit faster than the walking speed.  By examing these trajectories from the dataset, we observe that most of them encounter sudden congested situations or stay at one place for a long time which can not be learned from historical data. The histogram in Figure~\ref{fig:traj_hist} also proves that such trajectories are extremely rare.



 

\vspace{-0.15in}
\section{Conclusion}
\vspace{-0.05in}


In this paper, we present an end-to-end travel time estimation model, namely \textsc{\small DeepTravel}, which addresses the separate estimation problem of segment-based approaches and the non-training-based drawback of sub-path based approaches. We propose an unique feature extraction structure which takes multiple features into account. We also introduce the dual interval loss, which elegantly matches the characteristic of BiLSTM with that of trajectory data, to incorporate additional supervisory information naturally. We conduct experiments on real datasets to demonstrate the superiority of \textsc{\small DeepTravel}.

\clearpage
\balance{}
\bibliographystyle{named}
\bibliography{ijcai18_1}

\end{document}